\documentclass[12pt]{l4dc2022}

\usepackage{makecell}

\def\noti{\neg}
\newcommand{\PAR}[1]{\vskip2pt \noindent{\bf #1}}

\title[Deep Interactive Motion Prediction and Planning]{Deep Interactive Motion Prediction and Planning: \\ Playing Games with Motion Prediction Models}
\usepackage{times}

\author{%
 \Name{Jose L. Vazquez}$^1$ \Email{vjose@ethz.ch}\\
 \Name{Alexander Liniger}$^1$ \Email{alex.liniger@vision.ee.ethz.ch}\\
 \Name{Wilko Schwarting}$^2$ \Email{wilkos@csail.mit.edu}\\
 \Name{Daniela Rus}$^2$ \Email{rus@csail.mit.edu}\\
 \Name{Luc Van Gool}$^{1,3}$ \Email{vangool@vision.ee.ethz.ch}\\
 \addr $^1$CVL - ETH Zurich, $^2$CSAIL - MIT, $^3$PSI - KU Leuven
}

\begin{document}

\maketitle

\begin{abstract}%
   In most classical Autonomous Vehicle (AV) stacks, the prediction and planning layers are separated, limiting the planner to react to predictions that are not informed by the planned trajectory of the AV. This work presents a module that tightly couples these layers via a game-theoretic Model Predictive Controller (MPC) that uses a novel interactive multi-agent neural network policy as part of its predictive model. In our setting, the MPC planner considers all the surrounding agents by informing the multi-agent policy with the planned state sequence. Fundamental to the success of our method is the design of a novel multi-agent policy network that can steer a vehicle given the state of the surrounding agents and the map information. The policy network is trained implicitly with ground-truth observation data using backpropagation through time and a differentiable dynamics model to roll out the trajectory forward in time. Finally, we show that our multi-agent policy network learns to drive while interacting with the environment, and, when combined with the game-theoretic MPC planner, can successfully generate interactive behaviors. \url{https://sites.google.com/view/deep-interactive-predict-plan}
\end{abstract}


\section{Introduction}
Modern autonomy stacks, specifically those used for self-driving vehicles, consider prediction and planning different parts of the pipeline. However, this separation assumes a one-way interaction, where only the ego-vehicle is influenced by the other vehicles/agents. In reality, this assumption is not valid since the ego-vehicle also influences the other vehicles' decisions. Considering two-way interactions results in a game between all the agents, where all the vehicles plan simultaneously and find an equilibrium for the motion-planning game. Recent work developed methods to solve this problem \citep{Liniger2020ARacing, Schwarting2021StochasticSpace, LeCleach2022ALGAMES:Games}. However, these methods require reward functions for all the agents, which limits these methods to specific problems like autonomous racing or toy examples. To overcome this limitation, several groups proposed to learn the reward function using inverse optimal control or inverse Reinforcement Learning (RL) \citep{Sadigh2018PlanningState,Zeng2019,Behbahani2019, Abbeel2004ApprenticeshipLearning}. However, this kind of approach has two disadvantages, first, inverse RL problems are often ambiguous and only work well with linear function approximators, shallow neural networks \citep{Sadigh2018PlanningState}, or reward functions defined in the image space, which tend to be computationally and memory inefficient \citep{Zeng2019}. Second, after recovering the reward function, one still has to solve the computationally expensive game-theoretic planning problem \citep{Sadigh2018PlanningState, Schwarting2019SocialVehicles}. In our work, we propose an alternative route by directly learning a policy of the other agents, leveraging recent large-scale motion prediction datasets \citep{Houston2020,Ettinger2021LargeDataset}. Having access to a policy allows for best-response and leader-follower type algorithms to be used as a motion planner, drastically simplifying the interactive ego-motion planning problem.

\begin{figure}[t]
    \centering
    \includegraphics[width=0.95\textwidth]{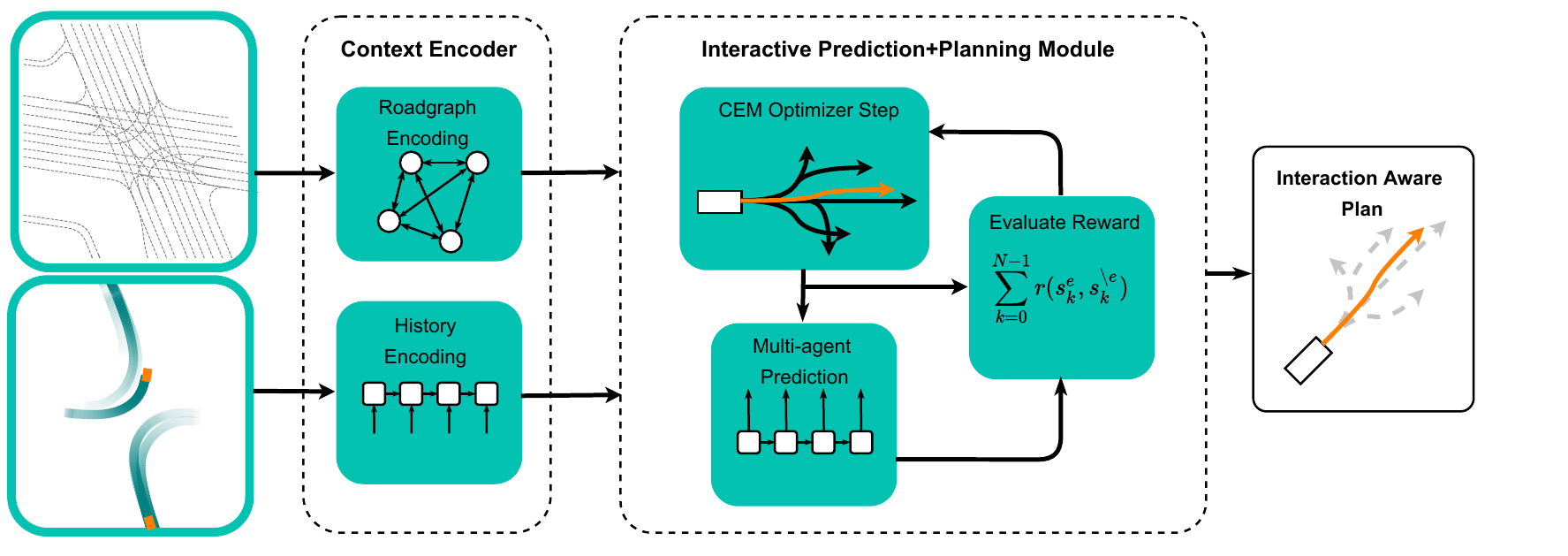}
    \caption[Joint Prediction-Planning System]{The joint prediction-planning module takes as inputs a past state buffer of all the vehicles in the scene and a HD-Map represented as a lane-graph. The Model Predictive Controller (MPC) inside the prediction-planning module uses the learned prediction module to internally simulate the multi-agent driving scenario.}
    \label{fig:system_diagram}
    \vspace{-0.7cm}
\end{figure}

By learning a policy, we can build on top of the ever-growing literature on deep motion prediction methods. However, as we will show, most of these approaches are highly optimized for accurate predictions but are not suited to reactively change to other vehicles in the prediction stage. Reactive prediction is a fundamental feature for our interactive planning approach, where the motion prediction policy of the other agents has to react to the proposed planned trajectory. Consequently, we propose using motion prediction tools such as road graph map integration and interaction layers but include them in an Interactive Multi-Agent Prediction (IMAP) policy. We further propose to train the policy in a model-based imitation learning approach. The core idea behind the IMAP policy is that at each step in the rollout/prediction, the model performs an interaction step both based on the recurrent state of the policy (intent) as well as the physical state of the agent. Additionally, the policy uses a map integration step where the relevant information is added from the HD-Map. Given all this information, the policy then computes actions for our differentiable dynamics model. Given the dynamics, we can roll out our policy and train it using an imitation loss on the physical states of the agents (position, heading, and velocity). We show that the policy can learn to predict and simulate a car on two large-scale motion prediction datasets, Lyft Level 5 \citep{Houston2020} and the Waymo Open Motion Dataset \citep{Ettinger2021LargeDataset}. In summary, our contributions are:

\begin{itemize}
    \setlength{\itemsep}{0.5pt}
    \setlength{\parskip}{0.5pt}
    \item An interactive prediction-planning module to simultaneously predict and plan trajectories, using game-theoretic planning.
    \item Formulating motion prediction as a policy learning problem, which we tackled using a novel model-based imitation learning approach.
    \item Design of an interactive motion prediction policy, which captures different types of interaction between the multiple agents as well as the map.

\end{itemize}

\section{Related Work}
 
\PAR{Prediction.} The motion prediction or forecasting problem focuses on regressing the vehicles' future states conditioned on the historical states and possibly map information as a context variable. Recently, the prediction problem has been tackled extensively using deep neural networks \citep{Ivanovic2018GenerativeBehavior}, but also model-based approaches like \citep{Hu2019GenericBehaviors} are still used due to their interpretability and data efficiency. The availability of large-scale motion forecasting datasets is the driving force of deep learning models for prediction, but also the ability to encode interactive modeling directly in the network architecture as a form of inductive bias is fundamental. Modeling interactions between all the agents in the scene has been addressed by using multi-headed attention models \citep{Mercat2020Multi-HeadForecasting,Rella2021DecoderPrediction} or by using Graph Neural Networks (GNN) \citep{Liang2020, Gao2020VectorNet:Representation, Li2020EvolveGraph:Reasoning, Kipf2018NeuralSystems, Graber2020DynamicInference}. More closely related to our work in planning conditioned prediction models, we can find PiP \citep{Song2020PiP:Driving} where a Convolutional Neural Network (CNN) style architecture is used to encode the ground truth future of the vehicle. Although this model could react to a planned trajectory, it does not handle agent-to-agent interactions in the decoding stage. Trajectron++ \citep{Salzmann2020Trajectron++:Control} uses a similar approach to encode the ground truth future of the ego vehicle but using a Conditional Variational Autoencoder (CVAE) framework \citep{Doersch2016} which better captures the distribution of possible paths an agent could follow conditioned on the ego-future but also lacks agent-to-agent interactions in the decoding stage. Trajectron++ was used for joint prediction and planning for human-robot interactions in \citep{Nishimura2020Risk-sensitiveInteraction}. \citep{Liu2021DeepPlanning} presents a reactive prediction model that can be used for planning purposes, the approach uses pre-defined trajectory sets for each traffic participant, and it is trained via an augmented cross-entropy loss that penalizes collisions with other vehicles. Similar to our proposed method, TrafficSim \citep{Suo2021} directly learns a policy in the decoder stage with the use of a differentiable simulation/dynamics model and a differentiable collision loss. Our approach aims to learn reactive behavior in continuous state-action spaces purely via imitation learning and without pre-defining a desired behavior in the loss function. Furthermore, we use a single recurrent cell that aims to capture the short and long term interactive behavior. 

\PAR{Planning.} Planning in autonomous driving, especially in an imitation learning setting, is a difficult task, particularly due to different state distributions during training and evaluation (covariate-shift) \citep{Ross2011}.  ChauffeurNet \citep{Bansal2019ChauffeurNet:Worst} addresses this by utilizing a clever data-augmentation technique that re-generates a feasible trajectory subject to a perturbation in the initial state of the agent.
In the interactive planning literature, Q-learning has been used to longitudinally control a vehicle using discrete accelerations \citep{Leurent2019}. 
Game-theoretic trajectory optimization algorithms for non-cooperative scenarios have been addressed in \citep{Spica2020ARacing, Wang2021Game-TheoreticScenarios, Williams2018BestVehicles, Fridovich-Keil2020, Di2019NewtonsGames}, but they require us to solve a full trajectory optimization problem for the other agents, which itself requires a pre-defined reward function. Reward functions have been successfully learned using inverse RL in \citep{Sadigh2018PlanningState, Schwarting2019SocialVehicles, Peters2021InferringObservations, Mehr2021Maximum-EntropySolutions}, but their structure is often limited to simple parameter vectors or too big image-based cost functions like in \citep{Zeng2019}. Our approach bypasses the reward function learning and uses a learned imitative policy that implicitly encodes the reward of the agents.

\section{Interactive Motion Prediction}
To train our IMAP policy, we employ a reinforcement and imitation learning-inspired approach. Specifically, we roll out our policy using a physical agent dynamics model. Given the rolled out agent trajectories, we can train the policy using a backpropagation trough time (BPTT) policy learning approach with an observation imitation loss. Thus, we can train a policy that generates physical actions only having access to observations. In the following, we will elaborate on this process, first introducing our model-based imitation learning approach given an abstract interactive policy and second introducing the exact design of our IMAP policy.

\subsection{Model-Based Imitation Learning}\label{sec:MBIL}
Theoretically, motion prediction can be posed as an imitation learning from observations problem \citep{Torabi2019RecentObservation,Edwards2019ImitatingObservation}, where we have observations of all the agents, and the goal is to find a policy that replicates the agents' decisions. The challenge with imitation learning from observations is that the actions are not available, rendering most imitation learning approaches unsuitable. However, in the case of motion forecasting, we have physical models that can predict the agents' motions since the main uncertainty does not come from the dynamics of the agents but the decision process. Considering that a physical model allows us to generate rollouts we can train the policies similarly to model-based RL methods \citep{Hafner2019, Clavera2020}, given that the model is differentiable with respect to the state and actions.
To make this more concrete, let us first introduce our multi-agent notation. We use the subscript $k$ to refer to a time step, and superscripts $i = 1,...,I$ to refer to an agent, with $I$ the number of agents. Furthermore, we use superscript $\noti i$ to refer to all agents except agent $i$. Second, we use a unicycle model as our motion model, as it well explains the motion of cars and other traffic agents. Thus, each agent is described by a state $s^i = [X, Y, \cos(\varphi), \sin(\varphi), v]$, where $(X, Y)$ is the position ($p$), $\cos(\varphi), \sin(\varphi)$ is used as a smooth heading representation, and $v$ is the forward velocity. Furthermore, an agent has the actions $a = [\alpha, r]$, where $\alpha$ is the acceleration and $r$ the yaw rate. Hence, the agent's motion can be described by a discrete-time dynamical system of the form $s_{t+1}^i = f(s_t^i,a_t^i)$. Since the state $s$ of the unicycle does not contain all information, such as the driving style, we use a recurrent policy, which has further access to a hidden state $h^i \in \mathbb{R}^H$. Finally, all agents have access to a map representation $m$. Therefore, we can define an abstract multi-agent policy $a = \pi(s^i, h^i, s^{\noti i}, h^{\noti i},m)$. The policy gets as the input the physical ego state $s^i$ and hidden state $h^i$, the physical state and hidden state of all the other agents $(s^{\noti i}$,  $h^{\noti i})$, and finally the map representation $m$.
In consequence, we can formulate the following model-based policy learning problem, 
\begin{align}
    \min_{\pi(\cdot)} \; \; & \sum_{k=0}^N \sum_{i=1}^I \mathcal{L}(s_k^i, \hat{s}_k^i) \nonumber\\
    \text{s.t. }\; \;  & s_0^i = \hat{s}_0^i \nonumber\\
    & s_{k+1}^i = f(s_k^i, a_k^i) \nonumber\\
    & \{a_k^i, h_{k+1}^i\} = \pi(s^i_k, h^i_k, s^{\noti i}_k, h^{\noti i}_k,m) \nonumber\\
    & i = 1,...I, \quad k = 0,...,N \label{eq:MBIL}
\end{align}
where, $N$ is the training horizon, and $\mathcal{L}$ is an imitation cost function (in our case, the Huber loss) that penalizes differences between the state trajectory of the policy rollout and the ground truth  measurements. Note that we encode the history of all the agents using the same policy, but using ground truth states (teacher forcing), according to 
$\{a_k^i, h_{k+1}^i\} = \pi(\hat{s}^i_k, h^i_k, \hat{s}^{\noti i}_k, h^{\noti i}_k,m)$, with $i = 1,...I$, and $k = -K,...,0$ where $K$ is the history length, finally, the hidden state is initialized with zero $h_{-K}^i = \mathbf{0}$.
The resulting policy learning problem is fully differentiable since the forward model is differentiable. Thus, we can use a BPTT policy learning algorithm, which can efficiently learn with the use of the differentiable dynamics.

\subsection{Interactive Policy} \label{sec:IMAP}

Given our model-based imitation learning problem \eqref{eq:MBIL}, we have a method to train our IMAP policy. However, to achieve a good performance, the structure of the policy is fundamental. For our design, we recognize three fundamental interaction types that must be considered: first, long-term intention interaction, second, physical interactions, and finally, map interactions.

\PAR{Policy Structure.} As explained in the previous section, we use a recurrent policy structure, which we implement as a Gated Recurrent Unit (GRU) cell. The GRU cell takes the previous hidden state $h$ as the recurrent input and additionally receives the concatenation of our three interaction modules, which will be explained next. Given the updated hidden state, we use two fully connected layers to produce the actions for the physical model. The output of our policy is a squashed Gaussian, which we sample from, this guarantees actions in physically reasonable bounds. However, it requires reparametrization \citep{Doersch2016} of the sampled actions to train our policy.

\PAR{Intention Interaction.} The idea behind the intention interaction is to capture the long-term driving style, goal, and intention interaction between agents. We argue that these attributes are represented in the hidden state $h$ of the policy. Hence, the interaction module considers the hidden representation of all agents. Inspired by \citep{Mercat2020Multi-HeadForecasting}, we use a multi-headed dot-product attention mechanism \citep{Vaswani2017AttentionNeed} to model the interaction. Consequently, the embedding of an agent can be used in combination with linear layers to generate the queries, keys and values for the attention module. Finally, the output is separated into the individual agents and used within the policy as an input to the GRU cell.

\PAR{Physical Interaction.} The second type of interaction is considering interactions between the physical states $s$ of agents. This interaction layer is intended for short-term behaviors and allows to learn collision avoidance and following skills. We formulate this layer as a GNN since it allows us to directly include state difference information in terms of edge features which is fundamental for this task. In our GNN, node features encode the physical state of each agent as well as agent metadata such as size and type, and edge features the difference in $(x, y)$ coordinates. We consider a fully connected graph with self connections, do allow for a fast spread of information. We follow \citep{Braso2020LearningTracking} for the implementation details and use the mean as an aggregation function.

\PAR{Map Interaction.} The agents interact not only with each other but also with the environment/map. Thus, map interaction is a core building block of modern motion prediction networks. In our network, we use a VectorNet \citep{Gao2020VectorNet:Representation} based map encoder. The encoder uses a GNN on the polylines describing the map; note that we denote the graph of polylines as $m$. Given the encoded polylines, we use an attention mechanism to attend to the important parts of the map. We use a cross-attention mechanism where the queries for each agent are generated from the output of the physical interaction layer. This two-stage approach first lets the information be encoded into an embedded map graph, followed by a learnable extraction mechanism. Note that the approach is also computationally efficient since the map embedding is shared across all agents in the scene.

\begin{figure*}[t]
    \centering
    \includegraphics[width=0.95\textwidth]{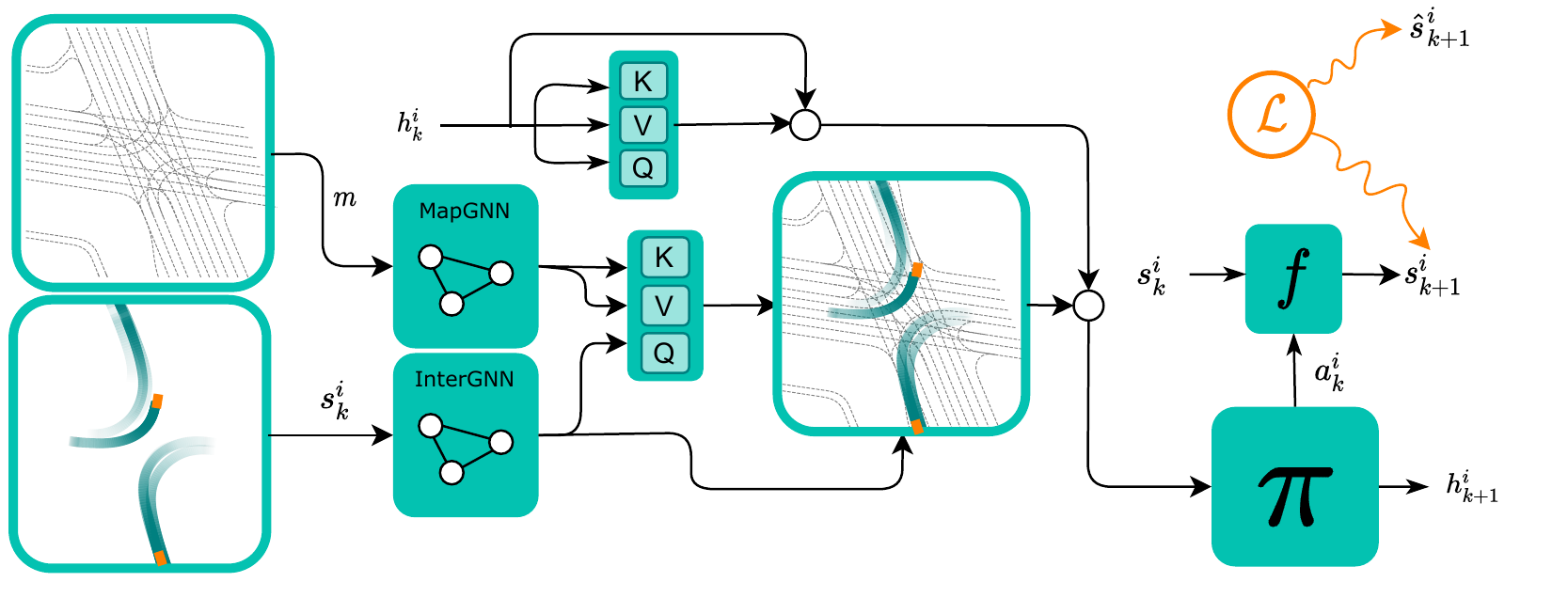}
    \caption[Interactive Policy]{The recurrent IMAP policy $\pi$ fuses the physical states $s_i^k$ and map information $m$ into a share embedding representation that is then used to to recursively control the dynamics model $f$. The single recurrent model is used for the encoding and decoding stage and is trained via BPTT using a direct loss on the states/observations $\mathcal{L}(\hat{s}_{k+1}^i, s_{k+1}^i)$. IMAP also uses the hidden embeddings to generate the Keys, Values and Queries (K, V, Q) used in the intention and map interaction network.}
    \label{fig:policy_diagram}
    \vspace{-0.7cm}
\end{figure*}

The complete architecture of the policy is shown in Fig.~\ref{fig:policy_diagram}, where we can see how the three interaction modules and the recurrent policy are set up. 

\PAR{Discussion.} The core idea behind the training scheme and the policy architecture is to train a model that can react to changing driving behaviors of other traffic agents. Accordingly, overemphasis on the encoding phase has to be avoided since otherwise, the agents only extrapolate, which can result in catastrophic failures \citep{Saadatnejad2021AreSocially-aware}. We achieve this by performing closed-loop training, see \eqref{eq:MBIL}. Thus, we do not learn to predict but learn to drive like the recorded agents. Note that the policy design is also essential, especially including the information about the other agents and the map at every step. This allows learning fundamental skills like collision avoidance and lane following. Hence, our method blurs the line between motion prediction and end-to-end driving, with our method being able to control traffic agents. One crucial difference is that our policy is not goal or route driven since we do not know the route of other agents. This is in contrast to end-to-end networks, which are normally conditioned on a route plan \citep{Zhang2021End-to-EndCoach, Bansal2019ChauffeurNet:Worst}. 

\section{Interactive Motion Planning}
Motion planning among other agents can be understood as a non-zero-sum game, where all agents plan trajectories considering their reward function. The reward function is agent-specific and captures the fundamentals of driving, such as collision avoidance, path following, and comfort. Solving this game has several drawbacks; first, the reward functions of other agents are generally unknown, and second, finding a Nash or other equilibrium of this game can be computationally demanding. 

Given our multi-agent policy, we can drastically simplify the problem since other agents' reward functions and even the best responses are directly embedded in the policy. Therefore, we can formulate the multi-agent motion planning problem by only optimizing over the action sequence $\mathbf{a^e} = [a^e_0, ..., a^e_{N-1}]$ of the ego agent by relying on the learned policy for all other agents. Accordingly, we only need to specify the ego reward function $r(s^e, s^{\noti e})$ and the interactive policy $\pi$. The resulting motion planning problem can be formulated as follows, 
\begin{align}
    \max_{\mathbf{a^e}} \; \; & \sum_{k=0}^{N-1} r(s_k^e, s_k^{\noti e}) \nonumber\\
    \text{s.t. }\; \; & s_0^e = \hat{s}_0^e, \; s_0^i = \hat{s}_0^i  \nonumber\\
    & s_{k+1}^e = f(s_k^e, a_k^e) \nonumber\\
    & s_{k+1}^i = f(s_k^i, \pi(s^i_k, h^i_k, s^{\noti i}_k, h^{\noti i}_k,m)) \nonumber\\
    & i = 1,...I, \quad k = 0,...,N \label{eq:MP_game}
\end{align}
where, identical to \eqref{eq:MBIL}, $f()$ is the discrete-time unicycle dynamics. Note that to simplify the notation, we omitted the recurrent output of the policy. Furthermore, $\noti i$ in this context still refers to all agents expect $i$, thus, the ego agent is considered inside the IMAP policy. 

We can solve this problem using derivative-free optimization solvers such as the Cross-Entropy Method (CEM). However, as usual in game theory, the order of play can have a drastic influence. Thus, in the following, we will propose two possible approaches, one resulting in a leader-follower style equilibrium and the second in a Nash style equilibrium. Note that these equilibria are for games in a trajectory space, similar to \citep{Liniger2020ARacing, Williams2018BestVehicles}, where each trajectory is interpreted as a strategy of the player.

Both approaches are based on best responses iterations, where agents iteratively update the strategy by the best possible actions given the current actions of the other agents. Our IMAP policy naturally reacts with a ``best" response to a other agents, even if the trajectory is fixed before hand. In the implementation we add the ego agent in the IMAP policy, and use the MPC trajectory with teacher forcing in the rollout. Note the we do not know the reward function of the IMAP policy.

\subsection{Iterative Leader-Follower MPC (ILF-MPC)}
Our first approach is a leader-follower setup inspired by the Stackelberg equilibrium \citep{Basar1998DynamicEdition}, where the leader is the ego agent. The ego agent computes a set of possible trajectories at each iteration using a CEM approach. The other agents compute their best response for all the ego trajectories by rolling out the IMAP policy. Given the best response trajectories, the reward of all the ego trajectories is computed, including collision penalties. The mean of the best trajectories (elites) is then used to initialize the next CEM iteration. Ideally, the CEM method is run until convergence. However, if the method is terminated early, the only drawback is that the ego trajectory could be further improved but the behavior of the other agents is still captured by the IMAP policy.

\subsection{Iterative Best-Response MPC (IBR-MPC)}
The leader-follower MPC proposed in the last section gives much power to the ego-agent, which potentially overestimates its influence. Therefore, we also propose an iterative best-response approach inspired by the Nash equilibrium \citep{Basar1998DynamicEdition}, since all players play a best response if the method converges, the resulting equilibrium is Nash. Therefore, we change the order of play and first compute trajectories for all agents with our IMAP policy. Given this starting point, the ego agent computes the best response with respect to the fixed agent trajectories using the CEM solver. Given this trajectory, all the other agents compute the best response using the IMAP policy with the ego trajectory fixed. This approach is repeated for several iterations or until convergence. Note that this approach needs fewer rollouts of the IMAP policy, but more CEM iterations since, at each outer-iteration, the ego agent needs to find the best trajectory requiring multiple CEM iterations.

\section{Results}\label{sec:Results}
The proposed interactive policy is tested in both the prediction and planning tasks. In the prediction task, the models are ablated against our re-implementation of the Argoverse 2019/2022 competition winner SAMPP \citep{Mercat2020Multi-HeadForecasting} based on standard single modal prediction metrics. Additionally, we show that adding the Non-Linear Least Square (NLLS) perturbation method proposed in \citep{Bansal2019ChauffeurNet:Worst} does not hurt the nominal performance of our models. In the planning task, we explore a lane merge scenario and show how IBP-MPC and ILF-MPC can plan a lane change behavior while also maximizing the margin to the approaching vehicle. Additionally we show how ILF-MPC can take advantage of having access to the reactive prediction model in the optimization step to plan highly interactive trajectories.

\subsection{Interactive Motion Prediction}

Essential to the success of our joint prediction-planning approach is the simultaneous prediction for all agents in the scene. We perform an ablation study to demonstrate that our information sharing mechanisms capture the underlying interactions of urban driving. For the ablation study we consider two of the most used metrics in trajectory prediction Final Displacement Error ($\text{ADE}$) and Average Displacement Error ($\text{ADE}$). Similar to other work in the field of trajectory prediction we perform inference with ego-centered scene information, but in contrast, we evaluate the model using the ego-agent metrics ($\text{ADE}^e$, $\text{FDE}^e$) as well as the rest of the agents' ($\text{ADE}^{\noti e}$, $\text{FDE}^{\noti e}$). Note that in our study, we do not consider multi-modal predictions but compare how different information sharing mechanisms can learn the underlying interactive behavior of the scene.

\PAR{Lyft level 5.} The Lyft level 5 motion prediction dataset \citep{Houston2020} consists of several full-length driving logs sampled at 10Hz. The logs contain information about the tracks of the perceived agents and the state of the ego vehicle. The level 5 dataset is not designed to output multi-agent motion sequences, therefore, both the train and validation datasets were preprocessed by subsampling the logs every 5 seconds and extracting all the motion sequences of the agents in the vicinity of the AV. Preprocessing the data allows for quicker training and evaluation on a single GPU. The models for the ablation study are trained only using tracks from vehicles (pedestrians and bicycles are removed from the dataset). The road graph information was resampled to have equidistantly spaced points for each polyline, similar to the polylines included in the Waymo dataset.

\PAR{Waymo.} The Waymo Open Motion Dataset (WOMD) \citep{Ettinger2021LargeDataset} consists of 9 second sequences sampled at 10Hz each sample contains multi-agent sequences of up to 8 actors that are probably interacting with each other. The dataset also contains a road graph per sample represented as a sequence of equidistantly sampled polylines. The models trained for the ablation studies on the Waymo dataset include the actor type (vehicle, pedestrian, cyclist) represented as a one-hot encoded embedding. Even though the dataset has sequences of 8s the models trained for the following ablation study are trained with a 3s horizon.

\PAR{Ablation Study.} All of the models were trained with a 3s prediction horizon, for the level 5 dataset, we used a 2s encoding buffer, and for the WOMD dataset its standard 1s encoding buffer. During training, with a probability $p=0.1$, the data is perturbed using random agent scene centering and random $SE(2)$ scene transformation with uniform distribution $(\Delta x, \Delta y) \sim \mathcal{U}[-10m, 10m]$ and $\Delta \varphi \sim \mathcal{U}[-0.8rad, 0.8rad]$. The perturbations are essential to mitigate the bias introduced by scene ego centering. The ablation study, performed on the level 5 and WOMD validation set, shows the same trends in both datasets, the SAMPP* baseline is outperformed by our proposed model with a differentiable unicycle propagation model and a stochastic policy. Conditioning SAMPP* on the ego ground truth trajectory SAMPP*(plan) similar to \citep{Salzmann2020Trajectron++:Control} does not improve the predictions of the other agents, showing that such an approach is not suited for our interactive planning. Once the interaction component is added to our method all the metrics improve, especially $\text{FDE}^{\noti e}$. The addition of the map encoder helps to improve the predictions of all agents but it particularly helps the ego predictions showing that the bias introduced due to scene centering is not fully mitigated via the data perturbations. Lastly, the NLLS perturbation needed for the planning module is added and we show that it does not hurt the nominal performance of our model. Note that we only train a single-mode model. As a result, our IMAP policy is not competitive in the multi-model evaluation metrics used in the WOMD-leaderboard.

\begin{table*}[t!]
\begin{center}
\caption{IMAP Policy Ablation Studies} 
\label{sample-table}
\renewcommand{\arraystretch}{1.1}
\begin{tabular*}{\textwidth}{l @{\extracolsep{\fill}} ccccccccc}
\\ \Xhline{3\arrayrulewidth}
 \textbf{Dataset} &  \multicolumn{4}{c}{Lyft @ 3s} & & \multicolumn{4}{c}{Waymo @ 3s}  \\ \cline{2-5} \cline{7-10} 
 \textbf{Metric} & $\text{ADE}^e$ & $\text{FDE}^e$ & $\text{ADE}^{\noti e}$ & $\text{FDE}^{\noti e}$ & & $\text{ADE}^e$ & $\text{FDE}^e$ & $\text{ADE}^{\noti e}$ & $\text{FDE}^{\noti e}$ \\ \Xhline{2\arrayrulewidth}
 SAMPP*          & 0.27 & 0.68 & 0.39 & 0.83 && 0.88 & 2.41 & 1.26 & 2.80  \\
 SAMPP*(plan)    & 0.06* & 0.08* & 0.41 & 0.84 && 0.31* & 0.43* & 1.49 & 3.06  \\[3pt] \Xhline{2\arrayrulewidth} 
 No Interact.       & 0.20 & 0.56 & 0.32 & 0.77  && 0.74 & 2.19 & 1.09 & 2.46 \\ 
 +Interact.          & 0.20 & 0.54 & 0.30 & 0.69  && 0.65 & 1.87 & 0.99 & 2.13 \\
 +Map      & 0.17 & 0.46 & 0.29 & 0.67  && 0.60 & 1.71 & 0.97 & 2.04 \\
 +NLLS           & 0.17 & 0.46 & 0.28 & 0.65  && 0.60 & 1.73 & 0.97 & 2.03 \\\Xhline{2\arrayrulewidth}
\end{tabular*}
\end{center}
\vspace{-0.5cm}
\end{table*}

\subsection{Interactive Motion Planning}
The motion planning component of the proposed architecture is tested in a qualitative fashion by analyzing a lane change scenario. Therefore, we take a scene from WOMD with seven agents and generate a plan for the ego agent, whereas the others are controlled by the WOMD trained IMAP policy.  
Lane changes are the most common type of interaction found in everyday driving and can be challenging in classical driving stacks. Often, lane changes require cooperation from cars on the merging lane. This interaction is not neglected in our approach compared to sequential prediction and planning. Thus, this scenario is used to explore the subtle differences between both proposed MPCs. Additionally, an adversarial reward term is used in the ILF-MPC method to showcase how privileged information about the reaction of other agents aids the optimization problem at hand.

\begin{figure}[t!]
    \centering
    \includegraphics[width=0.95\textwidth]{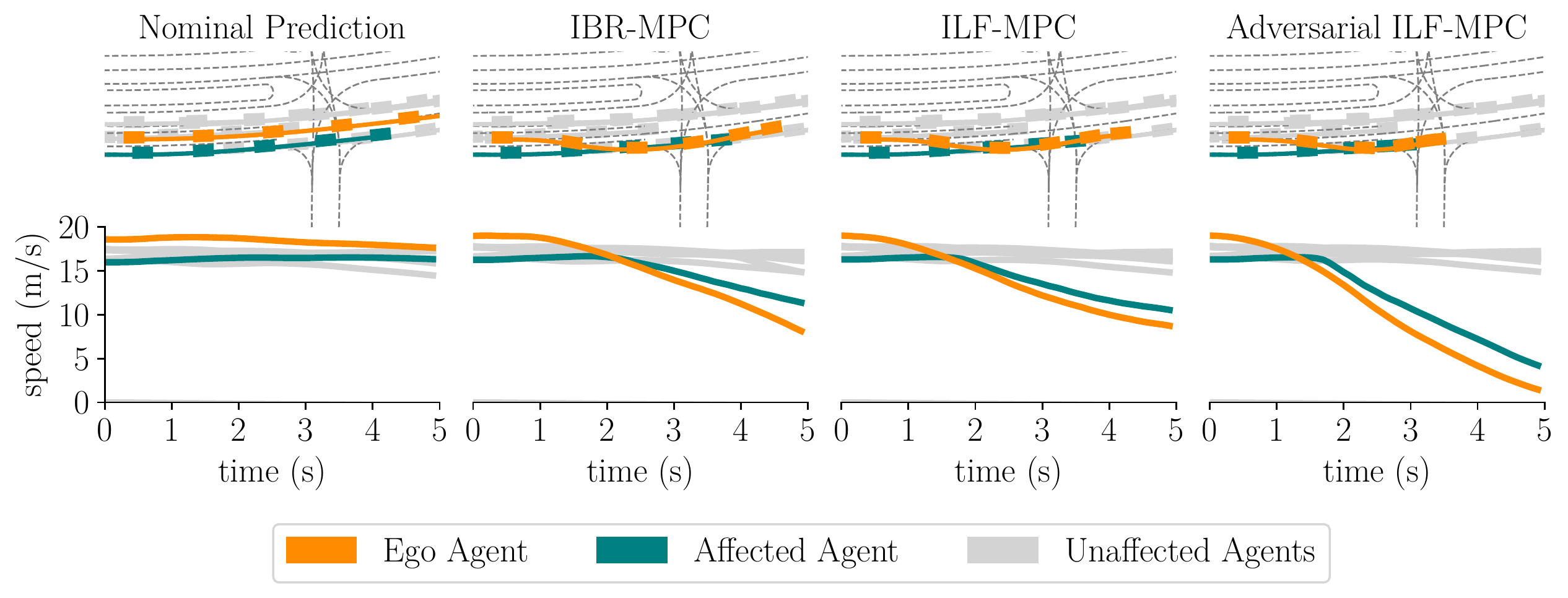}
    \caption[Adversarial reward fucntion]{Planned trajectories and velocities, form left to right, nominal model prediction for all agents from a WOMD sample, lane-change with IBR-MPC, lane-change with ILF-MPC, adversarial stopping example with ILF-MPC.}
    \label{fig:lane_change}
    \vspace{-0.7cm}
\end{figure}

\PAR{Best Response MPC vs. Leader Follower MPC.} To compare IBR-MPC and ILF-MPC, we solve the same lane-changing planning problem with a prediction horizon of 5s. The reward function uses a distance to target vehicle cost as a proxy for a collision penalty $ - 1 / ||p_k^e - p_k^{\noti e}||_2$, a terminal desired heading reward $- \text{anglediff}(\varphi_T^e - \varphi^*)$, a regularization term on the acceleration and yaw rate commands $- ||a_k^e||_2^2$ and finally the orthogonal distance of each point in the trajectory to the reference lane $-\text{proj}(p_k^e)$. The combined reward allows the optimization problem to find a solution that completes the lane change, but it also emphasises a safe distance to other vehicles. In Fig.~\ref{fig:lane_change} we can see that IBC-MPC and ILF-MPC have almost identical outcomes, but in our experiments we found that ILF-MPC tends to converge significantly faster than IBR-MPC using the same cost function and hyperparameters for the CEM optimizer. Next, we show how the ILF-MPC exploits its privileged information on how the agents would react. We introduce an adversarial reward term of the form $-v_k^a$, this reward incentivizes the ego agent to adversarially decrease the speed of the target agent. In Fig.~\ref{fig:lane_change} we can see that the ego agent plans to position itself in front of the target agent while braking, the policy of the affected agent performs a braking maneuver to avoid collision with the ego agent. Note that the optimizer is pushing the prediction model outside of the training data distribution, nonetheless, the policy reacts meaningfully and avoids a rear-end collision. The target agent does not come to a complete stop at the end. One could use a longer prediction horizon to overcome this issue or consider terminal constraints in the MPC. Note that we implemented the same interactive planners with SAMPP*(plan), but the MPC was not able to perform a lane change.

\section{Conclusion}
We proposed a novel interactive motion prediction and planning framework, where the motion prediction model and the motion planner play a game with each other. Therefore, we proposed a novel way to formulate the motion prediction problem as a policy learning problem. The policy is learned using model-based imitation learning, which, combined with our IMAP policy, allows us to learn a highly interaction-aware prediction model/policy. Given this model, we propose two interactive motion planners based on best response iterations. One is inspired by a leader-follower structure and the other by the Nash equilibrium. Finally, we showed simulation results in realistic driving situations where our deep interactive motion planning and prediction framework can perform challenging lane changes and even adversarial tasks such as stopping another car. Both our proposed methods can plan challenging interactive motions, correctly predicting the influence of the own action on the other agents. However, ILF-MPC can, by design, plan more interactive trajectories but it should be studied if human drivers behave according to the modeled leader-follower structure.

\bibliography{references.bib}

\section*{Appendix}

\section{Agent Dynamics}
We use a unicycle model for all the agents, since it models the non-holonomic behavior of cars and bicycles and well approximates the walking behavior of humans (normally no sideways walking). Thus, the continuous time dynamics of an agent is given by,
\begin{align*}
    \dot{X} &= v \cos(\varphi)\,,\\
    \dot{Y} &= v \sin(\varphi)\,,\\
    \dot{\varphi} &= r\,,\\
    \dot{v} &= \alpha\,.
\end{align*}
We use an Euler forward integrator to discretize the dynamics to our discrete time model $s_{k+1} = f(s_k, a_k)$. To enforce realistic motions we use input constraints of $r \in [-1, 1]$ and $\alpha \in [-5, 5]$, $\alpha$ is kept high to allow harsh braking maneuvers as seen in Fig. \ref{fig:lane_change}. In our experiments we also studied more complex models such as a kinematic bicycle model, and saw that it hinders the learning even if only cars are considered. We assume that this is due to the further non-linearities in the yaw dynamics. 
 
\section{Detailed Architecture of the IMAP model}

As described in Section \ref{sec:IMAP} the architecture of the IMAP network consists of different modules. In this section we describe in detail the architectural choices and the dimensionality of the inputs and outputs of all the components of the deep neural network. 

\begin{figure}[h]
    \centering
    \includegraphics[width=1.0\textwidth]{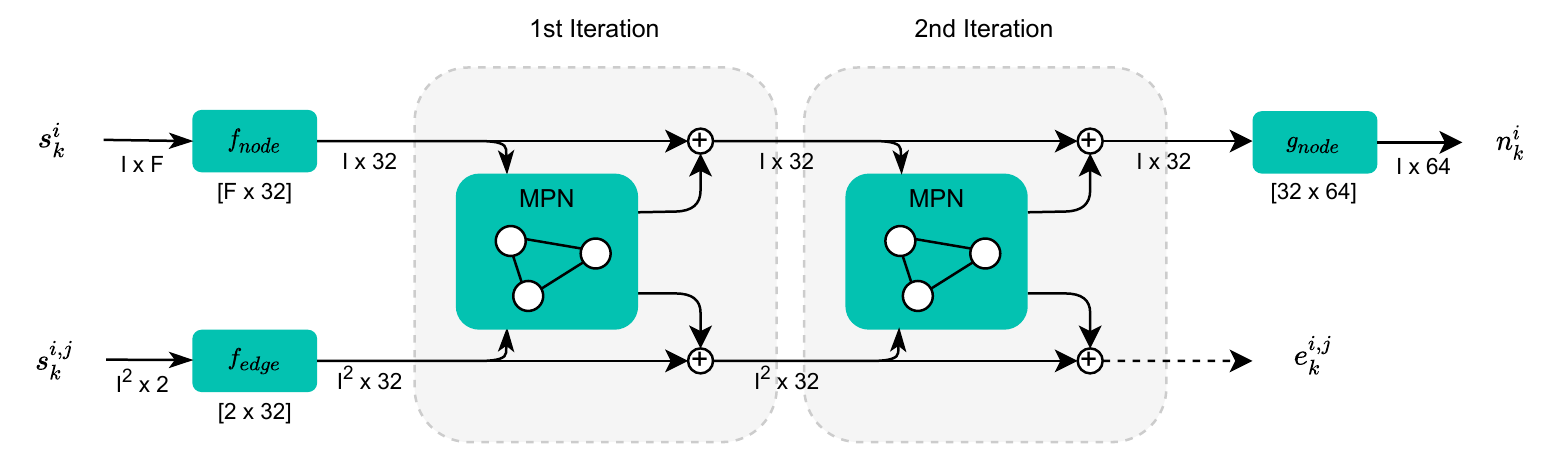}
    \caption[InterGNN]{Detailed architecture of InterGNN with damped iterations where $I$ is the number of vehicles in the scene and $F$ is the dimensionality of all the input features consisting of the vehicle state $s$ the vehicle extent (bounding box size) and a one hot encoding of the actor type.}
    \label{fig:inter_gnn}
\end{figure}

The information encoding stage is composed of several sub modules, the first is the InterGNN, which handles the \emph{physical interactions}. The module takes as an input the states of all agents described in Section \ref{sec:MBIL} and the relative positions $[\Delta x, \Delta y]$ between agents then it encodes them with a graph neural network, the Message Passing Netwotk (MPN) used inside the InterGNN is a GNN based on \citep{Braso2020LearningTracking} with slight modifications. Mainly the output of each MPN iteration is combined with the input using a convex mixing. This is inspired by damped fixed point iterations (Krasnoselski-Mann Iterations). 
Similar to the mixing in fixed point interactions and momentum in first order optimizers this should help the GNN to converge more reliably, which is also what we have seen in our experiments.
For the MPN we use a single linear layer followed by a ReLU non-linearity for both the node neural network $\mathcal{N}_n$ and the edge neural network $\mathcal{N}_e$. The MPN computes all the embeddings for all the agent nodes $n_i$ where $i \in \mathcal{I}$, and $\mathcal{I}$ is the set of all agents in the scene.
\begin{align*}
    e_{i, j}' &= \mathcal{N}_e([n_i, n_j , e_{i, j}]) \quad \forall j \in \mathcal{I} \\\
    n_i' &= \frac{1}{I} \sum_{j = 0}^{I}\mathcal{N}_n([n_j, e_{i, j}])
\end{align*}
Where the input-output dimensionality of $\mathcal{N}_n$ is $\text{[64, 32]}$ and the one from $\mathcal{N}_e$ is $\text{[96, 32]}$. In Fig.~\ref{fig:inter_gnn} the full InterGNN module architecture is visualized with the additional single layer neural networks used for encoding and decoding of the features. Empirically we found out that after two iterations the improvements in performance were insignificant.

After the embedding of the agents in the scene is computed the IntentAttention module captures the underlying hidden intention of the actors in the scene (intention interaction), this is motivated by classical partially observable Markov decision process literature and this information sharing module plus the GRU cell serve as a filtering step to learn the intention of all agents in the scene. The input of IntentAttention is the hidden state of the GRU cell from the previous step $h_{k}^i$ this embedding is used as the keys, queries and values for the multi-headed dot product attention. The attention module consists of 16 heads, and its input and output are of dimension 64.

\begin{figure}[ht]
    \centering
    \includegraphics[width=1.0\textwidth]{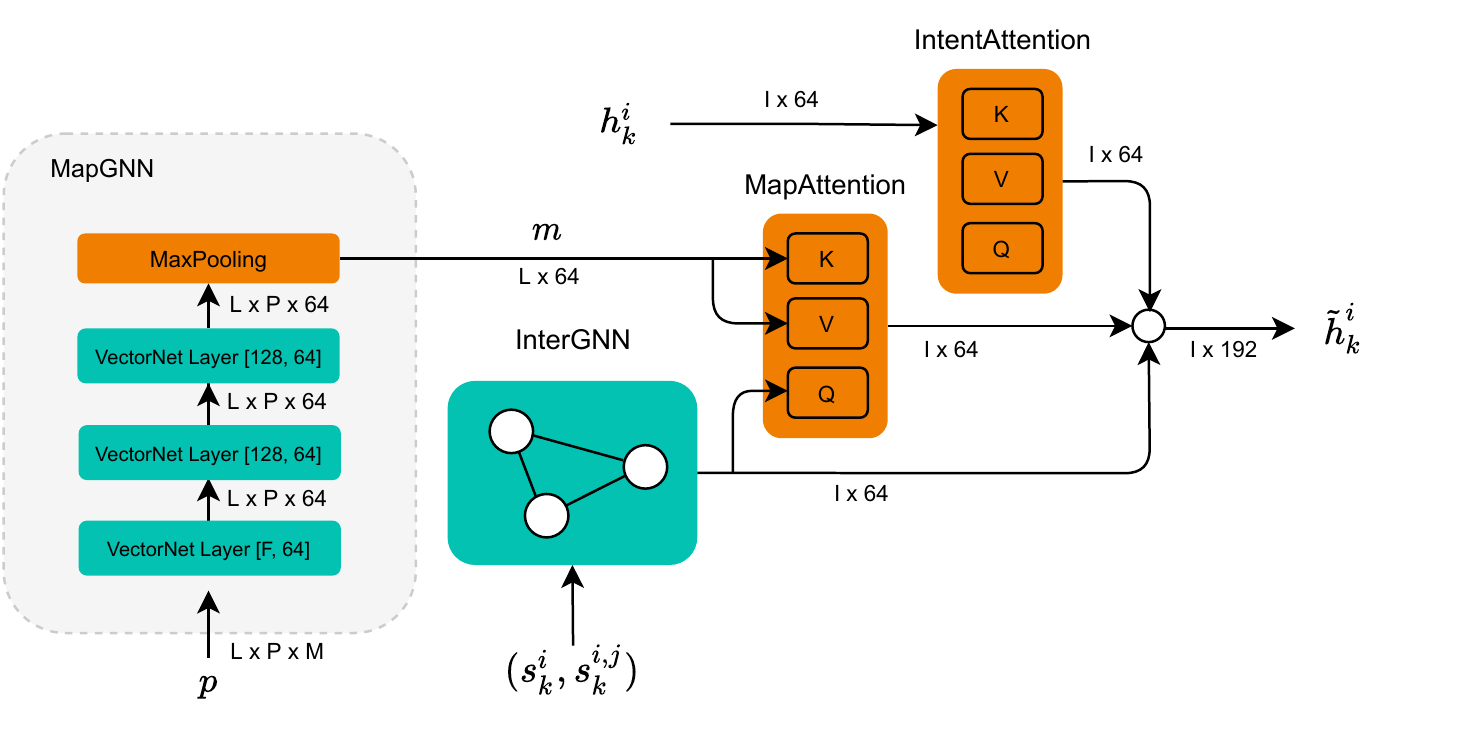}
    \caption[MapGNN]{Detailed architecture of encoding modules of the IMAP model. The output of the InterGNN module is used as the queries for the MapAttention module and then the inputs from InterGNN, IntentAttention and MapAttention are concatenated into $\tilde{h}_k^i$}
    \label{fig:map_gnn}
\end{figure}

The current state of the roadgraph is captured by the MapGNN, which is a roadgraph encoder network based on VectorNet \citep{Gao2020VectorNet:Representation}. In our implementation we excluded the map completion loss, therefore the network is only trained with the gradient information coming from the downstream modules. In the forward pass the network is only evaluated once, with the roadgraph centered at the zero position of the ego agent. The input roadgraph is represented as a set of $L$ polylines each one containing different amount of points but padded to be of size $P$. Each point can be represented with the following feature vector $p = [x, y, u_x, u_y, t, q]$ where $[x, y]$ is the position of that roadgraph element $[u_x, u_y]$ are the components of the unit vector that points to the next element in the polyline, $[t, q]$ are one hot encodings of the traffic light status and the polyline classification (e.g. crosswalk, double yellow) respectively. The map interaction is modeled using the MapAttention module, consisting of a a single headed dot product attention module. The attention module uses the InterGNN to decide on the query, thus allowing an position and interaction aware decision on what part of the roadgraph is fundamental for each agent. Finally, the MapAttention output is concatenated with the output of InterGNN and IntentAttention to the feature $\tilde{h}_k^i$. In Fig.~\ref{fig:map_gnn} we can see a detailed view of the InterGNN, MapGNN, and MapAttention with their respective input and output dimensions.

The combined embedding $\tilde{h}_k^i$ is used as the input to the GRU cell, with an input dimension of 192 and a hidden dimension of 64, we denote the GRU embedding by $h_{k+1}^i$. This output is then decoded using a two layer MLP with sizes [64, 64, 64, 2] and $\text{tanh}$ activation functions. The output of this MLP is a squashed normal distribution of the action space and it is trained via the reparametrization trick \citep{Doersch2016}.
 
\section{Training the IMAP network}

The IMAP network is trained with a Huber imitation loss on the target states, and during the history embedding phase we use teacher forcing without a loss. Empirically we found that the models trained without a loss on the speed targets have a more interactive behaviour and are able to generate better interacitve policies, mainly in terms of collision avoidance and close following of other cars. In contrast, models trained using a loss on speed targets sometimes fail to avoid frontal collisions and the interactive behaviour is less pronounced.

\section{ILF-MPC vs IBR-MPC}

In this section we explain in more detail the difference between ILF-MPC and IBR-MPC, to simplify the presentation we introduce the following notation conventions; $(\mathbf{s}, \mathbf{a})$ represent a state and action sequence for $k = 0, \dots , N$ and the matrices $(\mathbf{S}, \mathbf{A})$ represent batched sequences where each element in the first dimension is an individual state or action sequence. In Algorithm \ref{alg:ilfmpc} we show the algorithm for the ILF-MPC algorithm, note that the function $\text{RolloutPolicy}$ takes as input the full set of samples of the ego-agents state sequences $\mathbf{S}^e$. 

In contrast to ILF-MPC, IBR-MPC in Algorithm \ref{alg:ibrmpc} the rollout function $\text{RolloutEgoVehicle}$ uses single ego trajectory per best response iteration and then evaluates the rewards of the new sample ego trajectories against all the predicted agent trajectories that reacted to the previous ego-solution. Note that we make the assumption that a single CEM iteration will find a good solution because the policy is not being queried inside the CEM loop. The idea is that over multiple outer iterations the solution gets further improved, and our evaluations showed that this works well.

We choose a zero order optimization method like CEM over gradient optimization because it is more robust to local minima, which are a big issue in optimization based collision avoidance \citep{Zhang2021Optimization-BasedAvoidance}. Additionally CEM is very simple to implement and it does not require expensive gradient computation for the complex neural networks used in this work.

For our results in Section \ref{sec:Results} we used the following hyperparameters, ILF-MPC used 30 CEM iterations, and IBR-MPC 30 outer best response iterations and as said one CEM iteration, while both approaches use $M = 128$ action samples.

\SetKwInput{KwInput}{Input}                
\SetKwInput{KwOutput}{Output}              
\SetKwComment{Comment}{// }{ }
\SetKwComment{CommentEmpty}{}{}
\begin{algorithm}[!ht]
\DontPrintSemicolon
\KwInput{$f, r, \pi,s_0^e, h_0^e, s_0^{\noti e}, h_0^ {\noti e},  \mathbf{a}_{init}^e$}
\KwOutput{$\mathbf{s}^e, \mathbf{a}^e$}
$\mathbf{a}^e = \mathbf{a}_{init}^e$ \newline
\For{Every CEM iteration}{
    $\mathbf{A}^e \sim \mathcal{N}(\mathbf{a}^e)$ \Comment*[r]{Sample M action sequences} 
    $\mathbf{S}^e \gets \text{RolloutEgoVehicle}(s_0^e, \mathbf{A}^e, f) $ \CommentEmpty*[r]{}
    $(\mathbf{S}^{\noti e}, \mathbf{A}^{\noti e}) \gets \text{RolloutPolicy}(\mathbf{S}^e, s_0^{\noti e}, h_0^e, h_0^{\noti e}, f, \pi)$ \CommentEmpty*[r]{}
    $\hat{\mathbf{a}}^e \gets \textbf{k-max}(\text{EvaluateRewards}(\mathbf{S}^{e}, \mathbf{A}^{e}, \mathbf{S}^{\noti e}, \mathbf{A}^{\noti e}, r))$ \CommentEmpty*[r]{}
    $\mathbf{a}^e \gets \textbf{mean}(\hat{\mathbf{a}}^e)$ \Comment*[r]{Take the mean of elite sequences} 
}
$\mathbf{s}^e = \text{RolloutEgoVehicle}(s_0^e, \mathbf{a}^e, f) $ \newline
\caption{Iterative Leader-Follower MPC}
\label{alg:ilfmpc}
\end{algorithm}
\begin{algorithm}[!ht]
\DontPrintSemicolon
\KwInput{$f, r, \pi,s_0^e, h_0^e, s_0^{\noti e}, h_0^ {\noti e},  \mathbf{a}_{init}^e$}
\KwOutput{$\mathbf{s}^e, \mathbf{a}^e$}
$\mathbf{a}^e = \mathbf{a}_{init}^e$ \newline
\For{Every Best Response iteration}{
$\mathbf{s}^e \gets \text{RolloutEgoVehicle}(s_0^e, \mathbf{a}^e, f) $ \CommentEmpty*[r]{}
$(\mathbf{s}^{\noti e}, \mathbf{a}^{\noti e}) \gets \text{RolloutPolicy}(\mathbf{s}^e, s_0^{\noti e}, h_0^e, h_0^{\noti e}, f, \pi)  $ \CommentEmpty*[r]{}
$(\mathbf{S}^{\noti e}, \mathbf{A}^{\noti e}) \gets \text{Repeat}(\mathbf{s}^{\noti e}, \mathbf{a}^{\noti e})  $ \CommentEmpty*[r]{}
\For{Every CEM iteration}{
    $\mathbf{A}^e \sim \mathcal{N}(\mathbf{a}^e)$ \Comment*[r]{Sample M action sequences} 
    $\mathbf{S}^e \gets \text{RolloutEgoVehicle}(s_0^e, \mathbf{A}^e, f) $ \CommentEmpty*[r]{}
    $\hat{\mathbf{a}}^e \gets \textbf{k-max}(\text{EvaluateRewards}(\mathbf{S}^{e}, \mathbf{A}^{e}, \mathbf{S}^{\noti e}, \mathbf{A}^{\noti e}, r))$ \CommentEmpty*[r]{} 
    $\mathbf{a}^e \gets \textbf{mean}(\hat{\mathbf{a}}^e)$ \Comment*[r]{Take the mean of elite sequences} 
}
}
$\mathbf{s}^e = \text{RolloutEgoVehicle}(s_0^e, \mathbf{a}^e, f) $ \newline
\caption{Iterative Best-Response MPC}
\label{alg:ibrmpc}
\end{algorithm}

\newpage

\begin{figure}[h]
    \centering
    \includegraphics[width=0.85\textwidth]{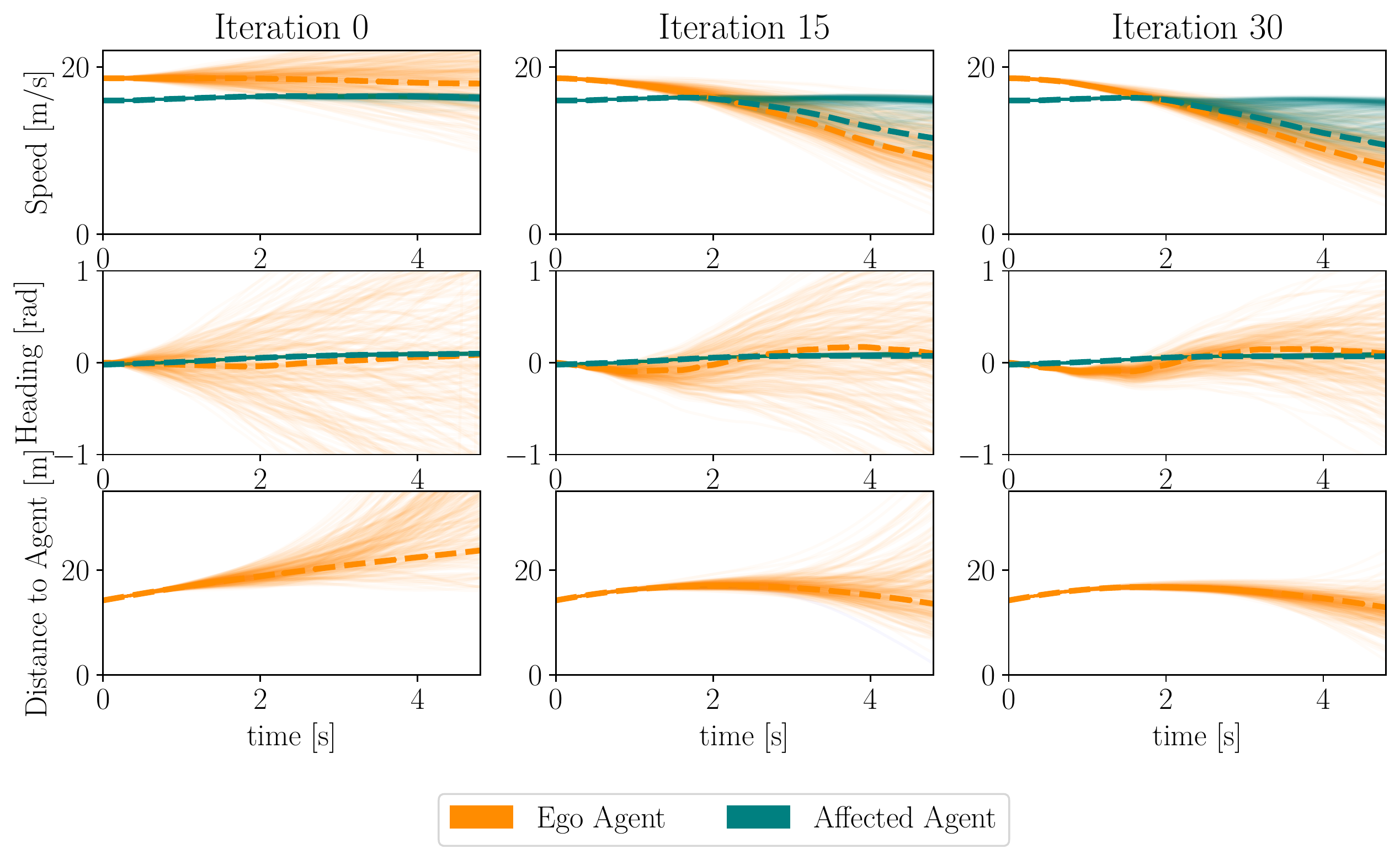}
    \caption[ilf iters]{Visualization of the samples trajectories generated per CEM iteration in the ILF-MPC.}
    \label{fig:iterations_ilf}
\end{figure}

\begin{figure}[h]
    \centering
    \includegraphics[width=0.85\textwidth]{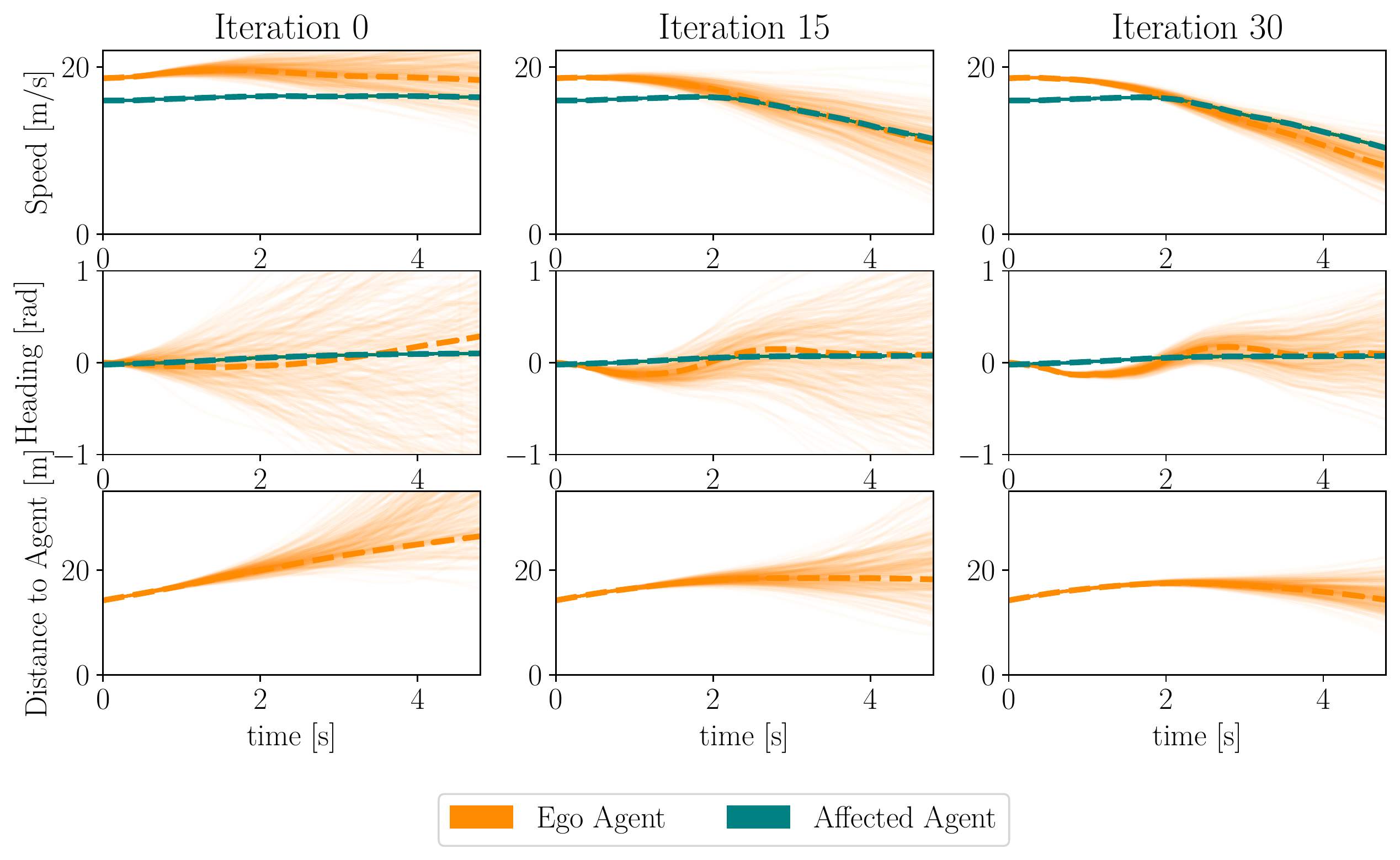}
    \caption[ibr iters]{Visualization of the samples trajectories generated per Best-Response iteration in the IBR-MPC. Note that a single CEM iteration per best response iteration is sufficient.}
    \label{fig:iterations_ibr}
\end{figure}

\begin{figure}[ht]
    \centering
    \includegraphics[width=1.0\textwidth]{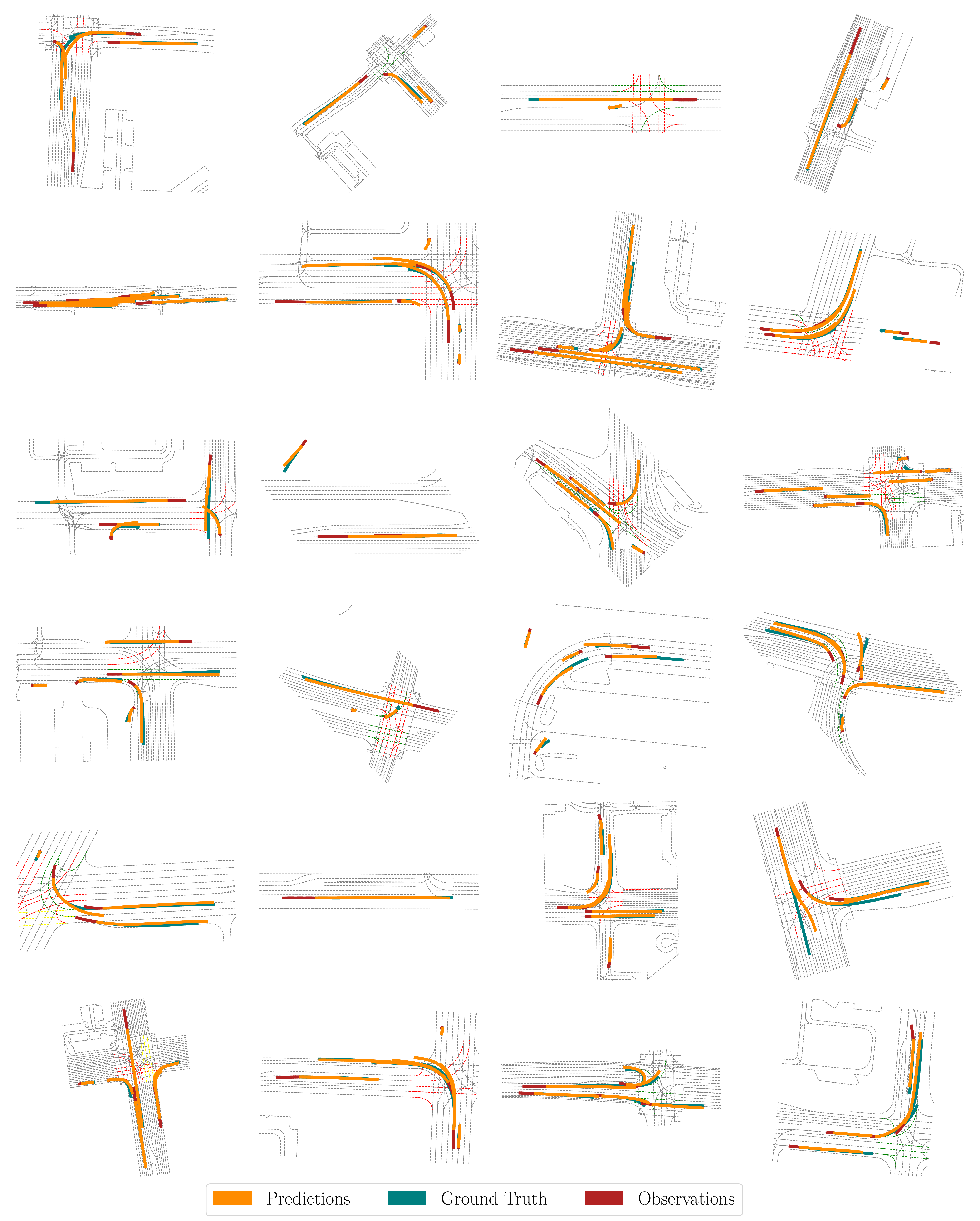}
    \caption[prediction examples]{Scenes where the IMAP policy correctly predicts the mode in the validation distribution, the policy very accurately performs lane following thanks to the VectorNet inspired map encoder.}
    \label{fig:prediction_examples}
\end{figure}

\begin{figure}[ht]
    \centering
    \includegraphics[width=1.0\textwidth]{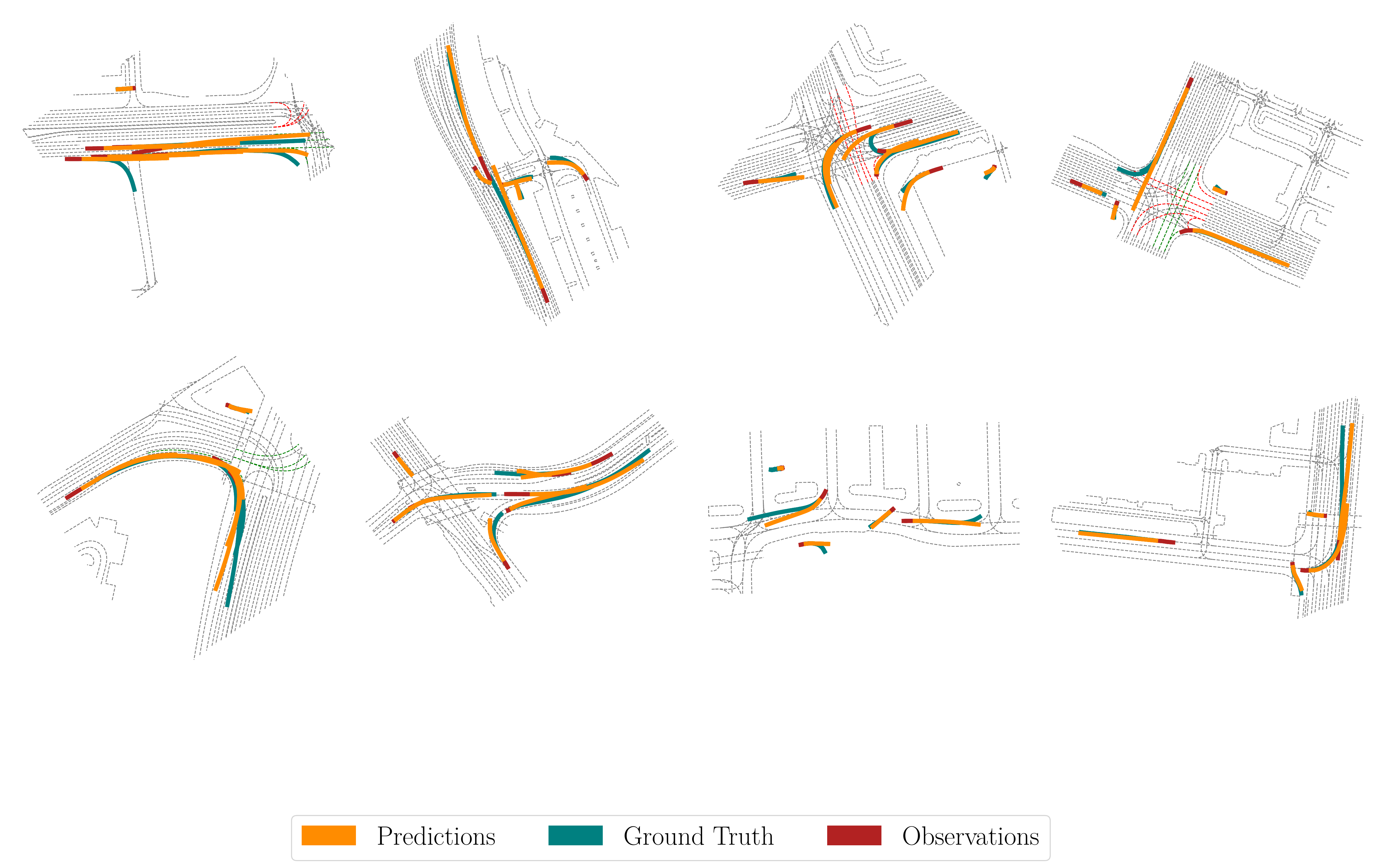}
    \caption[prediction bad examples]{Examples where the uni-modal IMAP model fails to accurately predict the correct mode in the validation distribution. }
    \label{fig:prediction_badexamples}
\end{figure}

\end{document}